\documentclass{article}

\PassOptionsToPackage{numbers, compress}{natbib}

\usepackage[preprint]{neurips_2023}

\usepackage[utf8]{inputenc} 
\usepackage[T1]{fontenc}    
\usepackage{hyperref}       
\usepackage{url}            
\usepackage{booktabs}       
\usepackage{amsfonts}       
\usepackage{nicefrac}       
\usepackage{microtype}      
\usepackage{xcolor}         
\usepackage{graphicx}       
\usepackage{subcaption}     
\usepackage{amsmath}        
\usepackage{natbib}         
\usepackage{amsthm}
\theoremstyle{definition}
\newtheorem{definition}{Definition}
\newtheorem{example}{Example}

\title{The Weighted M\"obius Score:\\A Unified Framework for Feature Attribution}

\author{%
  Yifan Jiang 
  \\
  Department of Linguistics\\
  University of Washington\\
  Seattle, WA 98195 \\
  \texttt{yfjiang@uw.edu} \\
  \And
  Shane Steinert-Threlkeld \\
  Department of Linguistics \\
  University of Washington \\
  Seattle, WA 98195 \\
  \texttt{shanest@uw.edu} \\
}

\begin{document}

\maketitle

\begin{abstract}
Feature attribution aims to explain the reasoning behind a black-box model's prediction by identifying the impact of each feature on the prediction. 
Recent work has extended feature attribution to interactions between multiple features.
However, the lack of a unified framework has led to a proliferation of methods that are often not directly comparable.
This paper introduces a parameterized attribution framework---the Weighted M\"obius Score---and (i) shows that many different attribution methods for both individual features and feature interactions are special cases and (ii) identifies some new methods.
By studying the vector space of attribution methods, our framework utilizes standard linear algebra tools and provides interpretations in various fields, including cooperative game theory and causal mediation analysis. 
We empirically demonstrate the framework's versatility and effectiveness by applying these attribution methods to feature interactions in sentiment analysis and chain-of-thought prompting.
\end{abstract}

\section{Introduction}

Explaining the predictions made by black-box machine learning models, such as neural networks, poses a significant challenge.
To address this challenge, feature attribution has become a popular approach, aimed at determining the impact of individual features on a model's prediction. 
However, the application of these attribution methods for explaining feature interactions, which play a critical role in many real-world scenarios, remains an open problem.
For example, in sentiment analysis, the interaction between the words ``not'' and ``bad'' results in a prediction that differs from what either word alone would produce. 
The capability to encode such interactions is believed to be the reason for the success of neural networks \citep{Goodfellow-et-al-2016} which highlights the need for a unified framework for feature attribution that can be applied to both individual features and feature interactions.

Recently, there has been a growing interest in extending attribution methods to feature interactions. 
Several methods have been proposed \citep{Tsang2017DetectingSI, Sundararajan2019TheST,Janizek2020ExplainingEA,Tsang2020HowDT}, but they are often not directly comparable due to their different assumptions. 
Empirical studies have produced conflicting results, and it is challenging to determine which method is superior, as the results may depend on the specific task and model being used. 
Furthermore, ground truth attributions are often not available for real-world tasks, making it difficult to compare these methods empirically.

This paper presents a unified framework---the \emph{Weighted M\"obius Score}---for model-agnostic feature attribution for both individual features and feature interactions (Section~\ref{sec:method}). 
Our framework situates feature attribution methods within a vector space, which is then analyzed using standard linear algebraic tools.
Our framework also has a natural interpretation in terms of cooperative game theory and causal mediation analysis, providing a unified perspective for understanding existing attribution methods and developing new ones (Section~\ref{sec:interpretation_and_connection_to_existing_work}).

Our contributions include:
(1) A unified framework for model-agnostic local feature attribution, based on linear algebra, that can be applied to both individual features and feature interactions.
(2) A theoretical analysis that bridges concepts from feature attribution, cooperative game theory, and causal mediation analysis.
(3) An empirical demonstration of the framework's versatility and effectiveness on real-world tasks such as sentiment analysis and prompt engineering (Section~\ref{sec:applications}).

\section{The Weighted M\"obius Score}
\label{sec:method}

In this section, we outline the framework for local feature attribution methods, which are designed to explain the reasoning behind a model's prediction for a single input.
Our framework is model-agnostic, meaning that it is applicable to any black-box model and does not rely on any assumptions about the model's architecture, training procedure, or mathematical properties.

\paragraph{Notations}
We denote the model as a function $f: \mathcal{X} \rightarrow \mathcal{Y}$, where $\mathcal{X}$ denotes a $d$-dimensional vector space and $\mathcal{Y}$ represents the output space.
We denote the set of all features in $\mathcal{X}$ by $\mathcal{D} = \{1, 2, ..., d\}$. 
For any input $x \in \mathcal{X}$ and any subset of features $S \subseteq \mathcal{D}$, counterfactual inputs $x_{\setminus S}$ can be constructed by removing the features in $S$ from $x$.
A thorough review of feature removal techniques can be found in \citet{Covert2020ExplainingBR}. 

\begin{definition}[Local Attribution Method] \label{def:local_attribution}
Given a model $f$ and an input $x$, a local attribution method $A$ is a function $A: \mathcal{P}(\mathcal{D}) \rightarrow \mathbb{R}$, where $\mathcal{P}(\mathcal{D})$ is the power set of $\mathcal{D}$. 
Some attribution methods only consider individual features and do not consider subsets with cardinality greater than one. 
These methods can be seen as a special case of $A$, where $A(S)=0$ for all $S$ with $|S|>1$. 
\end{definition}

\begin{definition}[Vector Space of Local Attribution Methods] \label{def:space_of_local_attribution}
The space of local attribution methods forms a vector space $\mathcal{A} = \mathbb{R}^{\mathcal{P}(\mathcal{D})}$, where $\mathbb{R}^{\mathcal{P}(\mathcal{D})}$ is the space of functions mapping from $\mathcal{P}(\mathcal{D})$ to $\mathbb{R}$.
The space has point-wise addition and scalar multiplication: $(A + B)(S) := A(S) + B(S)$ and $(cA)(S) := c \cdot A(S)$ for each subset $S \subseteq \mathcal{D}$.
\end{definition}
The space has dimension $2^d$ with a natural basis $\{\mathbf{1}_{S}: S \subseteq \mathcal{D}\}$, where $\mathbf{1}_{S}$ is the function that assigns 1 to $S$ and 0 to all other subsets.
This representation enables the application of standard linear algebra tools to analyze local attribution methods. 
The Zeta transform and its inverse, the M\"obius Transform, are examples of such tools \citep{Stanley2011EnumerativeCV}.

\begin{definition}[Zeta Transform and M\"obius Transform on $\mathcal{A}$] \label{def:zeta_and_mobius_transform}
The Zeta transform and the M\"obius transform are linear operators defined on function spaces with a partially ordered set domain. 
As such, they can be defined on the space of local attribution methods $\mathcal{A}$. Specifically, the Zeta transform $\zeta$ of a local attribution method $A$ is defined as $\zeta(A)(S) = \sum_{T \subseteq S} A(T)$.
The M\"obius transform $\mu$, on the other hand, is defined as $\mu(A)(S) = \sum_{T \subseteq S} (-1)^{|S| - |T|} A(T)$.
Here, $|T|$ represents the cardinality of the set $T$, and $(-1)^{|S| - |T|}$ is known as the M\"obius function. 
The Zeta and M\"obius transforms are inverses of each other, meaning that $\mu(\zeta(A)) = A$ and $\zeta(\mu(A)) = A$. 
These transforms offer a powerful framework for exploring the properties of local attribution methods.
\end{definition}

\begin{definition}[Feature Isolation Score] \label{def:model_discretization_method}
Let $f$ be a model with output space $\mathbb{R}$. \footnote{We can map any non-real output space $\mathcal{Y}$ to $\mathbb{R}$ via an appropriate transformation.}  Then the feature isolation score can be defined as
\[
A_f(S) = f(x_{\setminus \overline{S}}) - f(x_{\setminus \mathcal{D}})
\]
\end{definition}

This score evaluates the significance of a subset of features $S$ by comparing the predictions of $f$ on two counterfactual inputs: the first with the features outside $S$ removed, and the second with all features removed. 
Due to its simplicity and ease of implementation, the feature isolation score serves as a useful starting point for the development of more advanced attribution techniques.
Similar concepts have been explored in the literature, such as subset extension in \citet{Covert2020ExplainingBR}.

\begin{definition}[M\"obius Score] \label{def:mobius_score}
Given the feature isolation score $A_{f}$, the M\"obius Score $A_{\mu(f)}$ is defined as the M\"obius transform of $A_{f}$, i.e.:
\[
A_{\mu(f)}(S) = \mu(A_{f})(S)
\]
Alternatively, the M\"obius score can be recursively defined as:
\[
A_{\mu(f)}(S) = 
\begin{cases}
A_{f}(S) & \text{if } |S| = 0 \\
A_{f}(S) - \sum_{T \subset S} A_{\mu(f)}(T) & \text{if } |S| > 0
\end{cases}
\]
\end{definition}

The M\"obius score has the desirable \textit{efficiency} property, which means that the model's prediction for a given input can be completely decomposed into the sum of the M\"obius Scores of all feature subsets present in the input. 
As such, this leaves no contribution unattributed.

Furthermore, the M\"obius score satisfies a notion of \emph{identifiability}:
the M\"obius score for a feature subset identifies the highest-order interaction within that set.
For example, in a regression model with interaction terms, this means that the M\"obius score of each subset of features is equal to the corresponding terms in the model.

\begin{example}[Polynomial Model] \label{ex:polynomial_regression_model}
Consider a polynomial model $y = \beta_0 + \beta_1x_1 + \beta_2x_2 + \beta_3x_1^2 + \beta_4x_2^2 + \beta_5x_1x_2$. 
The M\"obius score of each subset of features is then as follows:
\[
A_{\mu(y)}(S) =
\begin{cases}
0 & \text{if } S = \emptyset \\
\beta_1x_1 + \beta_3x_1^2 & \text{if } S = \{1\} \\
\beta_2x_2 + \beta_4x_2^2 & \text{if } S = \{2\} \\
\beta_5x_1x_2 & \text{if } S = \{1,2\} \\
\end{cases}
\]
A more sophisticated example is the Taylor polynomial of a $d$-th continuous differentiable model $f$ around a baseline input $0$.
The M\"obius score of each subset of features is given by:
\[
A_{\mu(\text{Taylor}(f))}(S) = \sum_{I \in \mathcal{I}_{S}} \frac{D_{I}f(0)}{I!}x^I
\]
Here, $\mathcal{I}_{S}$ denotes the set of all multi-indices $I$ with $I_i \geq 1$ for all $i \in S$ and $I_i = 0$ for all $i \notin S$.
With this notation, $D_{I}f(0) = \frac{\partial^{|I|}f(0)}{\partial x_1^{I_1} \dots \partial x_d^{I_d}}$, $I! = \prod_{i \in S} I_i!$ and $x^I = \prod_{i \in S} x_i^{I_i}$.
\end{example}

\begin{definition}[Weighted M\"obius Score]
Given a weight function $\mathbf{w}: \mathcal{P}(\mathcal{D}) \times \mathcal{P}(\mathcal{D}) \rightarrow \mathbb{R}$, the \emph{weighted M\"obius score} is defined by
\[
A_\mathbf{w}(S) = \sum_{T \subseteq \mathcal{D}} \mathbf{w}(S,T)A_{\mu(f)}(T)
\]
We say that $\mathbf{w}$ (and the corresponding $A_\mathbf{w}$) is faithful just in case $\mathbf{w}(S,T) = 0$ if $S \cap T = \emptyset$. 
\end{definition}

\begin{definition}[Faithful Local Attribution Method]
A local attribution method $A$ is \emph{faithful} just in case $A = A_\mathbf{w}$ for some faithful $\mathbf{w}$.
\end{definition}

Intuitively, a faithful attribution method only attributes importance to relevant features and not to irrelevant ones. 
The set of all faithful local attribution methods is a subspace of $\mathcal{A}$, and each element in it is characterized by a weight function $\mathbf{w}$, which can also be interpreted as a linear operator on $\mathcal{A}$.
This definition allows for an analysis of existing attribution methods within a unified framework: we will now show that many existing attribution methods are faithful, differing only in the choice of $\mathbf{w}$. 

\section{Interpretation and Connection to Existing Work} \label{sec:interpretation_and_connection_to_existing_work}

In this section, we show that many existing attribution methods can be seen as instances of the faithful weighted M\"obius score, with different weight functions $\mathbf{w}$.  
We first focus on a family of methods inspired by cooperative game theory and then on causal mediation analysis.
A summary of the attribution methods discussed in this section (as well as the basic M\"obius score from Definition~\ref{def:mobius_score}) can be found in Table~\ref{tab:summary_of_attribution_methods}.
Proofs of the results in this section can be found in Appendix~\ref{sec:proofs}.

\begin{table}[htbp]
\centering
\caption{Summary of Attribution Methods. For all values not mentioned, $\mathbf{w}(S,T) = 0$.}
\begin{tabular}{@{}lll@{}}
\toprule
\textbf{Method} & \boldmath$\mathbf{w}(S,T)$ & \textbf{Order $^1$}  \\
\midrule
M\"obius Score & 1 if $S = T$ & Up to $|\mathcal{D}|$-order \\
Shapley Value & $\frac{1}{|T|}$ if $|S|=1$ and $S \subseteq T$ & First-Order  \\
Shapley Interaction Index & $\frac{1}{|T|-|S|+1}$ if $|S| \leq k$ and $S \subseteq T$ & Up to $k$-th Order  \\
Shapley-Taylor Interaction Index & $\begin{cases}
  1 & \text{if } |S| < k \text{ and } S = T \\
  \frac{1}{\binom{|T|}{k}} & \text{if } |S| = k \text{ and } S \subseteq T \\
  \end{cases}$ & Up to $k$-th Order \\
Pure Indirect Effect & 1 if $|S|=1$ and $S=T$ & First-Order \\
Total Indirect Effect & 1 if $|S|=1$ and $S \subseteq T$ & First-Order \\
Mediated Interaction Effect $^2$ & 1 if $|S| = 2$ or $3$ and $S = T$ & Second/Third-Order \\
ArchAttribute & 1 if $|S|=k$ and $T \subseteq S$ & $k$-th Order \\ 
\midrule
\multicolumn{3}{l}{$^1$``Order'' refers to the cardinality of the feature subset that the method can explain.} \\
\multicolumn{3}{l}{$^2$ A generalization to higher-order interactions has been proposed in this section.} \\
\bottomrule
\end{tabular}
\label{tab:summary_of_attribution_methods}
\end{table}

\subsection{Cooperative Game Theory}

\paragraph{Cooperative Game}
Given a model $f$ and an input $x$, we can model the attribution problem as a cooperative game $G = (N, v)$ where $N = \mathcal{D}$ is the set of players and $v = A_{f}$ is the payoff function.
In this game, each player $i \in N$ represents a feature in the input $x$ and each coalition $S \subseteq N$ represents a subset of features.
The goal of the game is for the players to form coalitions that maximize their joint payoff, or equivalently, to find a subset of features that maximizes the model output relative to the baseline input $x_{\setminus D}$.

\paragraph{Solution Concept}
The grand coalition $N$ is often assumed to yield the maximum payoff in the game $G$.
An allocation of the payoff of the grand coalition among the players is referred to as a solution concept of the game $G$.
The problem of feature attribution can be viewed as finding a solution concept of the game $G$ that satisfies certain desirable properties.
Consequently, each solution concept of the game $G$ can be seen as a local attribution method.

\paragraph{Harsanyi Dividend}
The Harsanyi dividend is a concept introduced by \citet{Harsanyi1958ABM} to analyze solution concepts.
It can be defined recursively as follows:
\[
d_v(S) = 
\begin{cases}
v(S) & \text{if } |S| = 0 \\
v(S) - \sum_{T \subset S} d_v(T) & \text{if } |S| > 0
\end{cases}
\]

This concept quantifies the surplus of a coalition $S$ that cannot be attributed to the surplus of its sub-coalitions.
It is important to note that \textbf{the Harsanyi dividend is equivalent to the M\"obius score when $v = A_{f}$}.
This equivalence provides a game-theoretic interpretation of the M\"obius score and supports the \textit{identifiability} property.

\paragraph{Shapley Value}
The Shapley value \citep{Shapley195317AV} is a solution concept known for its unique satisfaction of certain fairness axioms, making it widely used in feature attribution.
Many existing attribution methods can be viewed as approximations of the Shapley value \citep{Covert2020ExplainingBR}, including LIME \citep{Ribeiro2016WhySI} and SHAP \citep{Lundberg2017AUA}.
The Shapley value of a player $i$ can be defined using the Harsanyi dividend as follows \citep{Harsanyi1958ABM}: 
\[
\phi(i) = \sum_{T \subseteq N: i \in T} \frac{1}{|T|} d_v(T)
\]
The Shapley value allocates to each player a weighted sum of the Harsanyi dividend for all coalitions that include the player. 
The weight function, $\mathbf{w}(S,T)$, is given by $\frac{1}{|T|}$ if $\{i\} = S \subseteq T$ and $0$ otherwise.
This function suggests that the Harsanyi dividend for a coalition $T$ is divided equally among its players, providing a non-axiomatic rationale for the fairness of the Shapley value. 
Furthermore, because the Shapley value only considers coalitions that contain the player, it is \emph{faithful}.

\paragraph{Shapley (-Taylor) Interaction Indices}
Interaction indices, generalizing the Shapley value to higher-order interactions, assign a value to each coalition $S$ with a size up to $k$ (i.e., $I(S) = 0$ for $|S| > k$) while satisfying axioms analogous to the Shapley value. 
The Shapley interaction index \citep{Grabisch1999AnAA} and the Shapley-Taylor interaction index \citep{Sundararajan2019TheST} are two popular variants. 
The Shapley interaction index can be defined using the Harsanyi dividend as follows \citep{Grabisch1999AnAA}:
\[
I_{\text{SII}}(S) = \sum_{T \subseteq \mathcal{D}: S \subseteq T} \frac{1}{|T| - |S| + 1} d_v(T)
\]
The weight function, $\mathbf{w}(S,T)$, is given by $\frac{1}{|T| - |S| + 1}$ if $|S| \leq k$ and $S \subseteq T$, and $0$ otherwise. 
This implies that the Harsanyi dividend of $T$ is shared equally between $S$ (considered as a single player) and the remaining players in $T \setminus S$. 
The Shapley-Taylor interaction index can be also derived from the Harsanyi dividend as follows \citep{Sundararajan2019TheST, Hamilton2021AxiomaticEF}:
\[
I_{\text{STI}}(S) = 
\begin{cases}
d_v(S) & \text{if } |S| < k \\
\sum_{T \subseteq \mathcal{D}: S \subseteq T} \binom{|T|}{k}^{-1} d_v(T) & \text{if } |S| = k 
\end{cases}
\]
For $|S| < k$, $I_{\text{STI}}(S)$ is equivalent to the Harsanyi dividend of $S$, making the weight function $\mathbf{w}(S, T)$ equal to $1$ if $S = T$ and $0$ otherwise.
If $|S| = k$, the index represents the weighted average of the Harsanyi dividend for all coalitions containing $S$. 
In this case, the weight function $\mathbf{w}(S, T)$ is given by $\binom{|T|}{k}^{-1}$, which is the inverse of the number of sub-coalitions of size $k$ within $T$. 
Thus the Harsanyi dividend of a coalition $T$ is distributed evenly among all sub-coalitions of size $k$ within $T$. 

\subsection{Causal Mediation Analysis}

\paragraph{Causal Mediation Model}
A causal mediation model is a framework used to analyze the causal relationship between variables, specifically focusing on the mediating effect through certain intermediate variables, or mediators.
Given a model $f$ and an input $x$, we can model the attribution problem as a causal mediation model $(X, M, Y)$.
Within this framework, the treatment $X$ can be either the original input $x$ or the counterfactual input $x_{\setminus \mathcal{D}}$. 
The mediators $M$ correspond to the features in $\mathcal{D}$ that may be affected by the treatment, while the outcome $Y$ represents the model's prediction, given the treatment and the mediators. 
We assume that $M$ completely mediates the effect of $X$ on $Y$, meaning that $Y$ is independent of $X$ when conditioned on $M$.
Hence, $Y$ can be expressed as a function depending only on $M$: $Y(S) = f(x_{\setminus \overline{S}})$ for all $S \subseteq M$.
This function represents the model's prediction for the counterfactual input $x_{\setminus \overline{S}}$, where the features in $S$ have been removed.

\paragraph{Decomposition of the Total Effect}
The total effect (TE) measures the effect of a treatment on an outcome, without considering any mediators, which can be expressed as $Y(M) - Y(\emptyset)$ using the causal mediation model defined above.
It can be decomposed into direct and indirect effects \citep{Pearl2001DirectAI}, where direct effects are not mediated by any variables, and indirect effects are transmitted through one or more mediators.
There is no direct effect when the treatment effect is completely mediated by the mediators, such as in the model defined above.
Decomposing the total effect into indirect effects is useful in understanding the causal mechanisms of a treatment, and feature attribution can be seen as finding a decomposition of the total effect that highlights the importance of each mediator.

\paragraph{Mediated Interaction Effect} 
The mediated interaction effect (MI) is an indirect effect that quantifies the interaction effect between the mediator and the treatment or among multiple mediators. 
Initially proposed by \citet{VanderWeele2013ATD}, this concept has been extended to accommodate multiple mediators \citep{Bellavia2018DecompositionOT, Taguri2018CausalMA, Gao2022DecompositionOT}; but current definitions only apply to models with two or three mediators.
To address this limitation, we propose a general definition that can be applied to any number of mediators:
\[
\text{MI}(S) = \sum_{T \subseteq S} (-1)^{|S| - |T|} (Y(T) - Y(\emptyset))
\]
where $S \subseteq M$ represents a subset of mediators. 
Our definition is consistent with the original definition when $|S| \leq 3$.
Moreover, \textbf{the Mediated Interaction Effect is equivalent to the M\"obius Score when $Y(S) = f(x_{\setminus \overline{S}})$}.
In the next section, we will empirically demonstrate that MI, not having been used for feature attribution, can effectively identify interactions encoded in neural networks.

\paragraph{Pure and Total Indirect Effect}
The pure indirect effect (PIE) and the total indirect effect (TIE) are two of the most commonly used indirect effects in the literature of causal inference.
PIE has been extensively applied to the interpretation of neural networks \citep{Vig2020InvestigatingGB, Finlayson2021CausalAO}, while TIE has found its application in \citep{Ban2022TestingPL}. 
Both effects focus on the effect of a single mediator on the outcome $Y$.
The pure indirect effect can be derived from the mediated interaction effect as follows:
\[
\text{PIE}(i) = \text{MI}(\{i\}) 
\]
PIE measures the impact of mediator $i$ on the outcome $Y$ when all other mediators remain constant at their counterfactual values. 
The weight function, $\mathbf{w}(S, T) = 1$ if $|S| = 1$ and $S = T$, and $0$ otherwise. 
In contrast, the total indirect effect can be derived as follows:
\[
\text{TIE}(i) = \sum_{T \subseteq \mathcal{D}: i \in T} \text{MI}(T) 
\]
TIE measures the collective impact of all mediators in $\mathcal{D}$ on the outcome $Y$, transmitted through mediator $i$. 
The weight function for TIE, $\mathbf{w}(S, T) = 1$ if $|S| = 1$ and $S \subseteq T$, and $0$ otherwise.
The primary distinction between the two indirect effects lies in their treatment of interactions. 
PIE accounts for only the main effect of the mediator, while TIE considers both the main effect and all potential interactions between the mediator and other mediators. 

\subsection{Other Related Work}

\paragraph{Archipelago}
Archipelago \citep{Tsang2020HowDT} is a framework designed to extend attribution methods to feature interactions. 
It comprises two components: an interaction attribution measure, ArchAttribute, and an interaction detector, ArchDetect. 
Both components can be examined within our framework. 
ArchAttribute can be represented in terms of the M\"obius Score as follows:
\[
\phi(S) = \sum_{T \subseteq S} A_{\mu(f)}(T)
\]
This definition is essentially the Zeta transform of the M\"obius Score, which is equivalent to the Feature Isolation Score.
The weight function, $\mathbf{w}(S, T) = 1$ if $T \subseteq S$ and $0$ otherwise. 
ArchDetect, on the other hand, is designed to detect interactions between two features, and can be represented as $\overline{\omega}_{i, j} = \frac{1}{2h_i^2h_j^2}((\sum_{T \subseteq \mathcal{D}: i, j \in T}A_{\mu(f)}(T))^2 + (A_{\mu(f)}(\{i, j\}))^2)$.
Here, $h_i = |x_i - (x_{\setminus \mathcal{D}})_i|$ and $h_j = |x_j - (x_{\setminus \mathcal{D}})_j|$. 
ArchDetect differs from other methods in that it employs a non-linear function of the M"obius score, rather than a weighted version.
Despite this difference, it is clear from the formula that ArchDetect evaluates the combined impact of all potential interactions involving $i$ and $j$. 

\paragraph{Gradient-based Attribution Methods}
Gradient-based attribution methods have emerged as a popular class of techniques for explaining the predictions of machine learning models. These methods analyze the gradient of a model's output with respect to its input features and have been used to identify important features and interactions.
Several individual feature methods, including Integrated Gradients \citep{Sundararajan2019TheST}, SmoothGrad \citep{Smilkov2017SmoothGradRN}, and DeepLIFT \citep{Shrikumar2017LearningIF}, and extensions to feature interactions, such as Integrated Hessians \citep{Janizek2020ExplainingEA}, have been proposed.
However, \textbf{these methods depend on certain assumptions about the model's mathematical properties, which may not always hold.} 
Prior research \citep{Montavon2015ExplainingNC, Ancona2017TowardsBU} has investigated various approaches to unify gradient-based attribution methods.
More recently, \citet{Deng2023UnderstandingAU} proposed a framework that unifies the Harsanyi dividend and gradient-based attribution methods using Taylor expansion, which provides a potential direction to bridge our framework and gradient-based attribution methods.

\section{Applications} \label{sec:applications}
In this section, we first demonstrate the application of our framework to design new attribution methods for causal mediation analysis in sentiment analysis.
We then show how our framework can be used to compare existing attribution methods in a black-box prompt engineering setting.
\footnote{Code for all experiments is available at \url{https://github.com/1fanj/WMS}.}

\subsection{Designing New Attribution Methods: Sentiment Analysis}

\paragraph{Task Definition}
We focus on sentiment analysis, which classifies text into positive or negative sentiment.
Our goal is to understand the contribution of each word's hidden representations and their interactions within the model's decision-making process. 
We evaluate our methods using the SST-2 dataset \citep{Socher2013RecursiveDM} and employ BERT-large \citep{Devlin2019BERTPO} as our base model, obtaining fine-tuned weights \footnote{\url{https://huggingface.co/assemblyai/bert-large-uncased-sst2}} from the HuggingFace model hub \citep{Wolf2019HuggingFacesTS}.
We analyze the model's hidden representations using the causal mediation analysis framework, employing PIE as the individual attribution measure and the second-order MI as the interaction attribution measure.
To the best of our knowledge, this is the first application of MI in the feature attribution context.
We randomly sample 100 examples from the validation set and compute layer-wise attribution scores for each example.
We conducted the experiments on Google Colab using a single NVIDIA Tesla T4 GPU.

\begin{figure}[htbp]
\centering
\subfloat[]["This movie was not bad" (not from SST-2)]{\includegraphics[width=\textwidth]{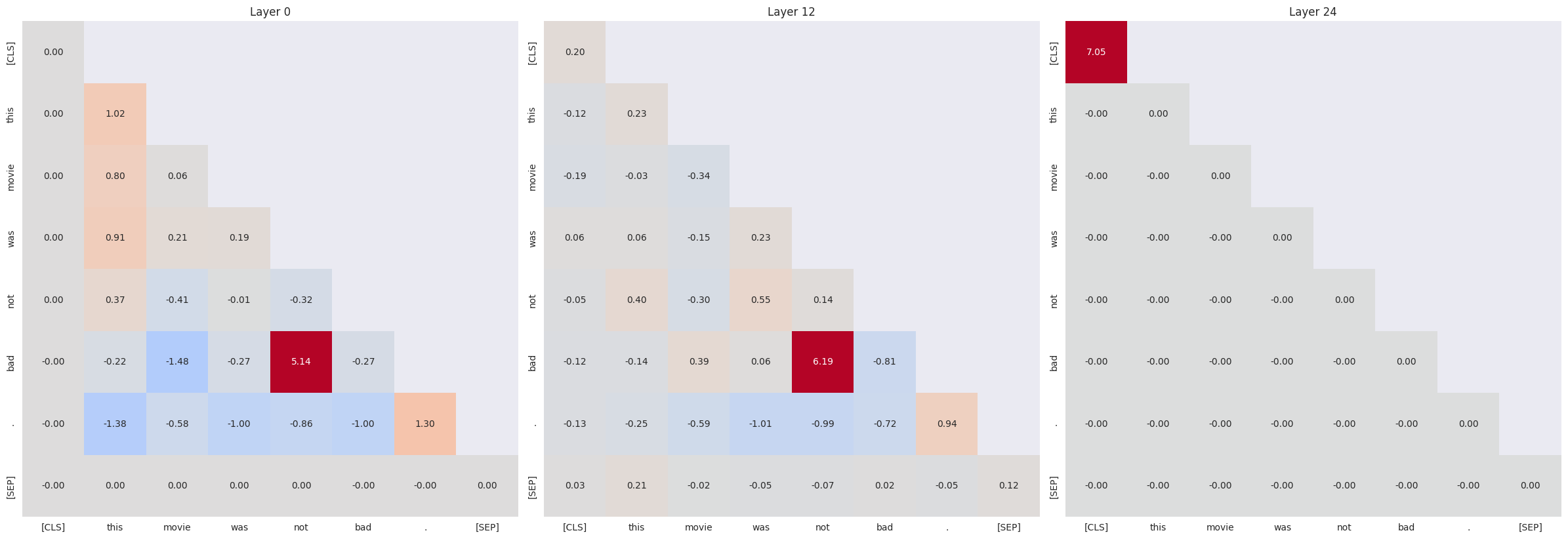}\label{fig:layerwise_mi_pie_1}} \\
\subfloat[]["The wild thornberrys movie is a jolly surprise"]{\includegraphics[width=\textwidth]{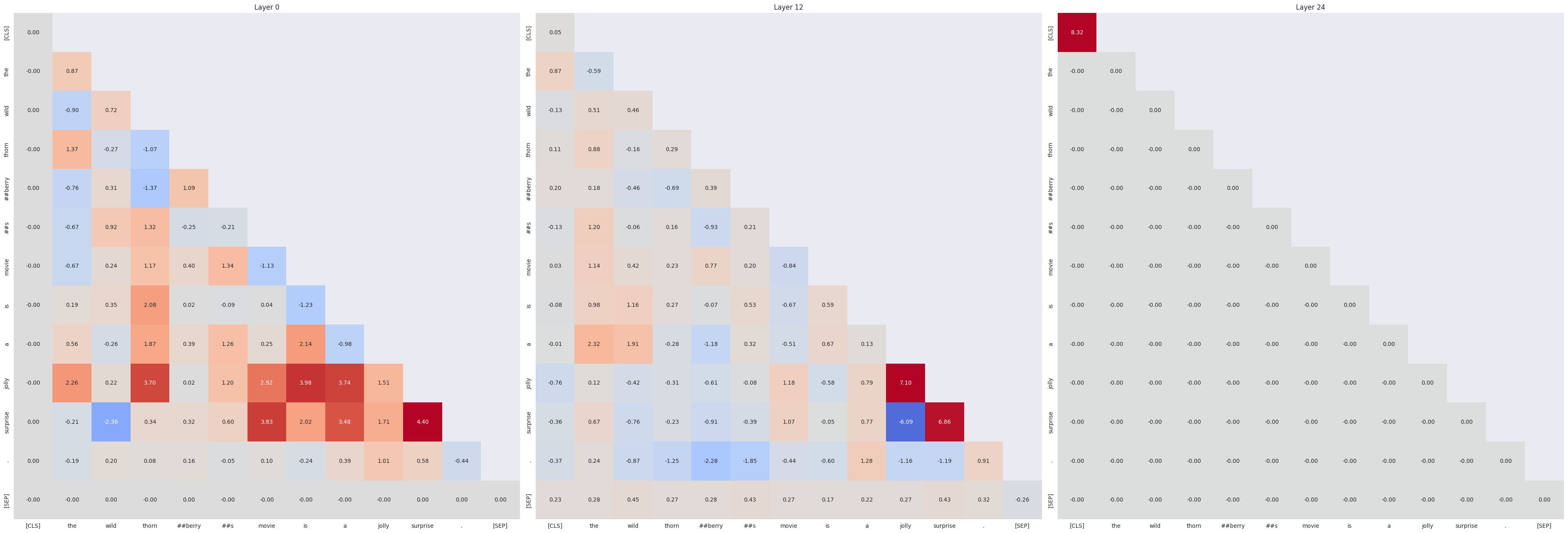}\label{fig:layerwise_mi_pie_2}}
\caption{Layer-wise MI and PIE scores for the BERT-large model on two example sentences.
Three layers are depicted in each graph: the first layer (Layer 0), the middle layer (Layer 12), and the final layer (Layer 24). 
The diagonal of each matrix represents the PIE scores, while the off-diagonal represents the MI scores. 
Colors represent the sign and magnitude of the attribution scores, with redder shades indicating more positive scores and bluer shades indicating more negative scores.}
\label{fig:layerwise_mi_pie}
\end{figure}
  
\paragraph{Layer-wise MI and PIE}
Figure \ref{fig:layerwise_mi_pie} illustrates layer-wise MI and PIE scores for two sentences. 
We observe stronger MI scores in lower layers and more pronounced PIE scores in higher layers, suggesting lower layers are more sensitive to word interactions, while higher layers focus on individual words. 
In the final layer, nearly all effects concentrate in the CLS token, which is used for the final classification decision.
We note that for both examples the CLS token has a high attribution to the positive sentiment, consistent with the model's prediction. 
In Figure \ref{fig:layerwise_mi_pie_1}, we notice a strong mediated interaction between ``not'' and ``bad'' with positive scores, despite both words having negative PIE scores, which is consistent with our expectations. 
Moreover, in Figure \ref{fig:layerwise_mi_pie_2}, we identify a strong interaction between ``jolly'' and ``surprise'' with negative scores, even though both words have positive PIE scores. 
This counter-intuitive phenomenon, referred to as ``saturation'' in \citet{Janizek2020ExplainingEA}, arises when interacting words share the same sentiment polarity as the model's prediction.

\begin{figure}[htbp]
\centering
\includegraphics[width=\textwidth]{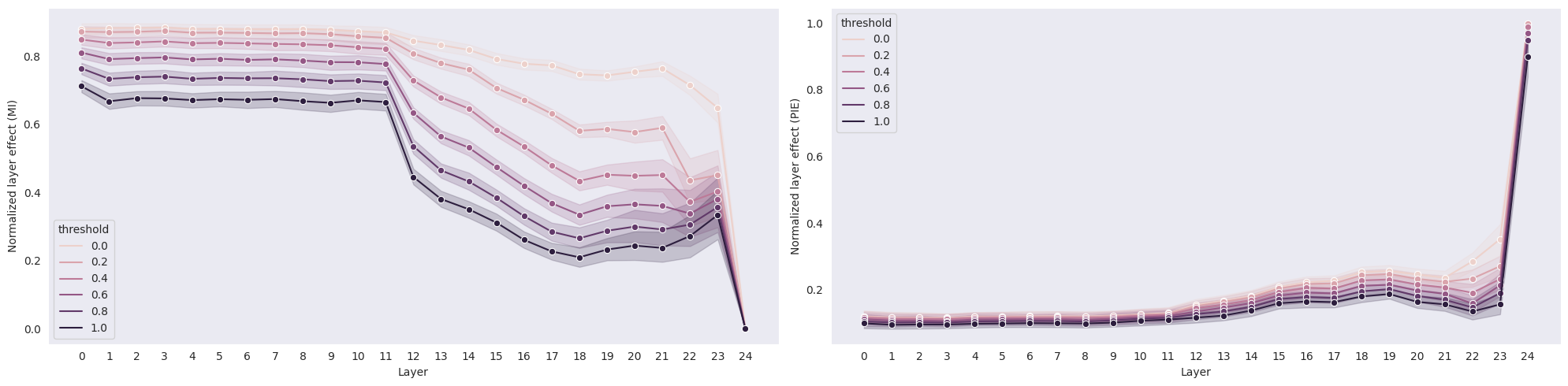}
\caption{Normalized layer effect for the BERT-large model across all 100 examples from the SST-2 dataset.
The $x$-axis represents the layer index, and the y-axis displays the normalized layer effect.
Shaded regions show 95\% confidence intervals.
Thresholds range from 0 to 1 in increments of 0.2.
Left plot: normalized MI; Right plot: normalized PIE.}
\label{fig:normalized_layer_effect}
\end{figure}

\paragraph{Normalized Layer Effect}
We calculate the normalized layer effect for MI and PIE, defined as the average proportion of the total magnitude (exceeding a threshold) of a specific type of effect to the total magnitude of all effects.
Figure \ref{fig:normalized_layer_effect} visualizes the normalized layer effect for MI and PIE. 
We find that normalized MI scores are higher in lower layers and gradually decrease from layer 12 to layer 24 until they reach zero, whereas normalized PIE scores exhibit the opposite trend. 
The trends are statistically significant, as indicated by the narrow confidence intervals. 
This finding further supports our previous observation on individual examples that the effects increasingly concentrate on individual tokens as the layers deepen.
These results demonstrate the effectiveness of the new attribution methods in revealing the different contributions and interactions of words in various layers of the model, which provides insights into the model's internal decision-making process, including the encoding and utilization of lexical relationships.

\subsection{Comparing Existing Attribution Methods: Prompt Engineering}

\paragraph{Task Definition}
We focus on prompt engineering, a task aiming to enhance a language model's performance on a downstream task by supplying prompts to the model.
One well-known method is Chain of Thought (CoT) \citep{Wei2022ChainOT}, which uses a sequence of demonstrations to guide the model's reasoning process.
Our objective is to understand the contribution of each sentence or word in the demonstrations to the model's performance.
We evaluate our methods on the last letter concatenation task proposed in \citet{Wei2022ChainOT}, in which the model concatenates the last letter of each word in a given name.  
OpenAI's ChatGPT API  (gpt-3.5-turbo) \citep{Brown2020LanguageMA, ChatGPT} \footnote{Although the results may not be replicable due to the model’s closed-source nature, the main focus of this experiment is the comparative analysis of attribution methods, which remains relevant and applicable.} is used to obtain the model's predictions with the temperature set to 0 to minimize randomness.
We select the first 100 examples from the dataset \footnote{\url{https://github.com/jasonwei20/chain-of-thought-prompting}} and compute the attribution scores on a one-shot CoT prompt.
We construct the input as follows:

\begin{itemize}
\item[\textbf{User}:] ``Take the last letters of the words in "Bill Gates" and concatenate them.''
\item[\textbf{Assistant}:] \{Demonstrations\} + ``The answer is ls''
\item[\textbf{User}:] ``\{Question\}''
\end{itemize}

where \{Demonstrations\} is a sequence of demonstrations which we vary in our experiments, and \{Question\} is the question we ask the model, e.g., ``Take the last letters of the words in "Waldo Schmidt" and concatenate them.''.
If the correct concatenation appears in the model's response, we consider the model's prediction to be correct.
We compute the M\"obius score and four additional attribution scores using the respective weight functions: the Shapley value, the Shapley interaction index, the total indirect effect, and ArchAttribute.

\begin{table}[htbp]
\centering
\caption{Sentence-level attribution scores for the last letter concatenation task. }
\begin{tabular}{l|c|c|c|c|c}
\toprule
\textbf{Sentences} & \textbf{M\"obius} & \textbf{Shapley} & \textbf{SII} & \textbf{TIE} & \textbf{ArchAttribute} \\
\midrule
\textbf{\#1} & 1.000 & 0.407 & 0.407 & 0.000 & 1.000 \\
\textbf{\#2} & 0.987 & 0.400 & 0.400 & 0.000 & 0.987 \\
\textbf{\#3} & 0.571 & 0.193 & 0.193 & 0.000 & 0.571 \\
\textbf{\#1, \#2} & -0.987 & 0.000 & -0.708 & 0.000 & 1.000 \\
\textbf{\#1, \#3} & -0.571 & 0.000 & -0.292 & 0.000& 1.000 \\
\textbf{\#2, \#3} & -0.558 & 0.000 & -0.279 & 0.000 & 1.000 \\
\textbf{\#1, \#2, \#3} & 0.558 & 0.000 & 0.558 & 0.000 & 1.000 \\
\bottomrule
\end{tabular}
\label{tab:sentence_level}
\end{table}
  
\paragraph{Sentence-level Attribution}
Table \ref{tab:sentence_level} presents the sentence-level attribution scores for the last letter concatenation task. 
The demonstrations consist of three sentences: `The last letter of ``Bill'' is ``l''.' (\#1), `The last letter of ``Gates'' is ``s''.' (\#2), and `Concatenating them is ``ls''.' (\#3).
Attribution scores are averaged across 77 out of 100 examples where the model's prediction is incorrect without the demonstrations. 
Our framework yields several insights into each method's behavior:
(1) The M\"obius score for all sentence pairs is negative, indicating the saturation phenomenon.
(2) TIE is entirely uninformative because it magnifies the interaction effects by attributing them to each involved sentence without any normalization.
(3) Both the Shapley value and the Shapley interaction index assign nearly equal importance to the first two sentences, as they distribute the interaction effects uniformly across the involved sentences.
(4) The Shapley interaction index attributes nearly equal importance to the pair \#1, \#2 and the pair \#2, \#3 because it considers the pair as a whole and then distributes the interaction effect uniformly across the involved sentences.
(5) The ArchAttribute score matches the model's accuracy, as it is exactly the feature isolation score, which is the difference between the model's outputs when only considering sentences of interest and when excluding the demonstrations entirely.
These insights demonstrate the usefulness of our framework in comparing attribution methods, thereby facilitating a deeper understanding of their strengths and weaknesses.

\begin{table}[htbp]
\centering
\caption{Phrase-level attribution scores for the last letter concatenation task. }
\begin{tabular}{l|c|c|c|c|c}
\toprule
\textbf{Phrases} & \textbf{M\"obius} & \textbf{Shapley} & \textbf{SII} & \textbf{TIE} & \textbf{ArchAttribute} \\
\midrule
\textbf{NP} & 0.605 & 0.206 & 0.206 & 0.000 & 0.605 \\
\textbf{PP} & 0.855 & 0.331 & 0.331 & 0.000 & 0.855 \\
\textbf{VP} & 1.000 & 0.450 & 0.450 & 0.000 & 1.000 \\
\textbf{NP, PP} & -0.566 & 0.000 & -0.276 & 0.000 & 0.895 \\
\textbf{NP, VP} & -0.618 & 0.000 & -0.329 & 0.000& 0.987 \\
\textbf{PP, VP} & -0.868 & 0.000 & -0.579 & 0.000 & 0.987 \\
\textbf{NP, PP, VP} & 0.579 & 0.000 & 0.579 & 0.000 & 0.987 \\
\bottomrule
\end{tabular}
\label{tab:phrase_level}
\end{table}

\paragraph{Phrase-level Attribution}
Table \ref{tab:phrase_level} presents the phrase-level attribution scores for the last letter concatenation task.
We compute the attribution scores for the noun phrase (NP), prepositional phrase (PP), and verb phrase (VP) in the first sentence of the demonstrations, which are `The last letter', `of ``Bill''', and `is ``l''', respectively.
Similarly, attribution scores are averaged across 76 \footnote{The number of examples differs from sentence-level attribution due to prediction randomness.} examples where the model fails without the demonstrations. 
Observations from Table \ref{tab:phrase_level} are similar to those from Table \ref{tab:sentence_level}.
Additionally, we notice that each individual phrase has a significant influence on the model's prediction, as evidenced by the positive M\"obius Scores assigned to each phrase and the negative scores assigned to each pair of phrases, indicating that the presence of key phrases alone can boost performance.
This suggests that the CoT prompt's effectiveness may not arise from the step-by-step reasoning process guiding the model, but instead from emphasizing key phrases that enhance its performance, which is consistent with the findings of \citet{Wang2022TowardsUC} that CoT prompts can be effective even with invalid demonstrations.

\section{Conclusion} \label{sec:conclusion}
In this paper, we propose a novel model-agnostic framework for understanding the behavior of local feature attribution methods.
Our framework introduces the \emph{weighted M\"obius score}, which is a principled measure for quantifying the interaction effects between features.
We show that this framework can be interpreted in various fields, including cooperative game theory and causal mediation analysis, thereby providing a unified view of feature attribution methods.
We demonstrate our framework's usefulness by designing a new attribution method tailored to causal mediation analysis and comparing various feature attribution methods in a fully black-box setting.
Our framework can be extended to other attribution methods and applications, which we leave for future exploration.
Improving computational efficiency, currently a bottleneck of our framework due to the exponential complexity of the M\"obius Score, is also a potential direction for future work.

\bibliographystyle{plainnat}
\bibliography{weighted_mobius_score}

\clearpage
\appendix

\section{Proofs} \label{sec:proofs}
In this appendix, we provide complete proofs for the results presented in Section \ref{sec:interpretation_and_connection_to_existing_work}.
It is important to note, however, that 
(i) proofs related to game-theoretic methods have been excluded as they are well-established in the literature and 
(ii) the definitions used in these proofs may differ from their original presentations in the literature due to the different notations used in this paper.

\paragraph{Pure Indirect Effect}
The pure indirect effect for a mediator $i$ can be represented as:
\[
\text{PIE}(i) = \text{MI}(\{i\})
\]

\begin{proof}
We start from the definition of the pure indirect effect:
\[
\text{PIE}(i) = Y(\{i\}) - Y(\emptyset)
\]
Adding and subtracting $Y(\emptyset)$ from the right-hand side leads to:
\[
\text{PIE}(i) = Y(\{i\}) - Y(\emptyset) - (Y(\emptyset) - Y(\emptyset)) = \text{MI}(\{i\})
\]
\end{proof}

\paragraph{Total Indirect Effect} \label{sec:proof_tie}
The total indirect effect for a mediator $i$ can be represented as:
\[
\text{TIE}(i) = \sum_{T \subseteq \mathcal{D}: i \in T} \text{MI}(T)
\]

\begin{proof}
We start from the definition of the total indirect effect:
\[
\text{TIE}(i) = Y(\mathcal{D}) - Y(\mathcal{D} \setminus \{i\})
\]
Under the assumption that $Y(S) = f(x_{\setminus \overline{S}})$, we add and subtract $f(x_{\setminus \mathcal{D}})$:
\[
\text{TIE}(i) = f(x) - f(x_{\setminus \{i\}}) = f(x) - f(x_{\setminus \mathcal{D}}) - (f(x_{\setminus \{i\}}) - f(x_{\setminus \mathcal{D}}))
\]
Now, we substitute $A_f(S) = f(x_{\setminus \overline{S}}) - f(x_{\setminus \mathcal{D}}) = \zeta(A_{\mu(f)})(S)$:
\[
\text{TIE}(i) = A_f(\mathcal{D}) - A_f(\mathcal{D} \setminus \{i\}) = \zeta(A_{\mu(f)})(\mathcal{D}) - \zeta(A_{\mu(f)})(\mathcal{D} \setminus \{i\}) 
\]
Expanding the definition of $\zeta$ and applying the inclusion-exclusion principle yields:
\[
\text{TIE}(i) = \sum_{T \subseteq \mathcal{D}} A_{\mu(f)}(T) - \sum_{T \subseteq \mathcal{D}: i \notin T} A_{\mu(f)}(T) = \sum_{T \subseteq \mathcal{D}: i \in T} A_{\mu(f)}(T) = \sum_{T \subseteq \mathcal{D}: i \in T} \text{MI}(T)
\]
\end{proof}

\paragraph{ArchAttribute} 
The ArchAttribute score for a set of features $S$ can be represented as:
\[
\phi(S) = \sum_{T \subseteq S} A_{\mu(f)}(T)
\]

\begin{proof}
We start from the definition of ArchAttribute:
\[
\phi(S) = f(x_{\setminus \overline{S}}) - f(x_{\setminus \mathcal{D}})
\]
Substituting $A_f(S) = f(x_{\setminus \overline{S}}) - f(x_{\setminus \mathcal{D}}) = \zeta(A_{\mu(f)})(S)$ and expanding the definition of $\zeta$ yields:
\[
\phi(S) = \zeta(A_{\mu(f)})(S) = \sum_{T \subseteq S} A_{\mu(f)}(T)
\]
\end{proof}

\paragraph{ArchDetect} \label{sec:proof_archdetect}
The ArchDetect score for a pair of features $(i, j)$ can be represented as:
\[
\overline{\omega}_{i, j} = \frac{1}{2h_i^2h_j^2}(\sum_{T \subseteq \mathcal{D}: i, j \in T}A_{\mu(f)}(T))^2 + (A_\mu(f)(\{i, j\}))^2  \\
\]

\begin{proof}
We start from the definition of ArchDetect:
\[
\overline{\omega}_{i, j} = \frac{1}{2}((\frac{1}{h_ih_j}(f(x) - f(x_{\setminus \{i\}}) - f(x_{\setminus \{j\}}) + f(x_{\setminus \{i, j\}})))^2 + (\frac{1}{h_ih_j}(f(x_{\setminus \overline{\{i, j\}}}) - f(x_{\setminus \overline{\{j\}}}) - f(x_{\setminus \overline{\{i\}}}) + f(x_{\setminus \mathcal{D}})))^2)
\]
Substituting $A_f(S) = f(x_{\setminus \overline{S}}) - f(x_{\setminus \mathcal{D}})$, the first addend inside the outermost parentheses becomes:
\[
\frac{1}{h_i^2h_j^2}(A_f(\mathcal{D}) - A_f(\mathcal{D} \setminus \{i\}) - A_f(\mathcal{D} \setminus \{j\}) + A_f(\mathcal{D} \setminus \{i, j\}))^2
\]
Similarly, the second addend becomes:
\[
\frac{1}{h_i^2h_j^2}(A_f(\{i, j\}) - A_f(\{i\}) - A_f(\{j\}))^2
\]
Rewriting both addends in terms of $\zeta(A_\mu(f))$ and expanding the definition of $\zeta$, we obtain:
\[
\overline{\omega}_{i, j} = \frac{1}{2h_i^2h_j^2}(\sum_{T \subseteq \mathcal{D}}A_{\mu(f)}(T) - \sum_{T \subseteq \mathcal{D}: i \notin T}A_{\mu(f)}(T) - \sum_{T \subseteq \mathcal{D}: j \notin T}A_{\mu(f)}(T) + \sum_{T \subseteq \mathcal{D}: i, j \notin T}A_{\mu(f)}(T))^2 + (A_\mu(f)(\{i, j\}))^2  \\
\]
Applying the inclusion-exclusion principle to the first addend yields:
\[
\overline{\omega}_{i, j} = \frac{1}{2h_i^2h_j^2}(\sum_{T \subseteq \mathcal{D}: i, j \in T}A_{\mu(f)}(T))^2 + (A_\mu(f)(\{i, j\}))^2  \\
\]
\end{proof}

\end{document}